\def\year{2018}\relax
\documentclass[letterpaper]{article} 
\usepackage{aaai18}
\usepackage{courier}
\usepackage{graphicx}
\usepackage{helvet}
\usepackage{times}
\usepackage{url}
\usepackage{amsthm}
\usepackage{stfloats}
\usepackage{amsmath}
\usepackage{amssymb}
\usepackage{comment}
\usepackage{float}
\usepackage{textcomp}
\usepackage{footnote}
\usepackage{multirow}
\usepackage{booktabs}
\usepackage{makecell}
\usepackage{tabularx}
\usepackage[table]{xcolor}
\usepackage{pgfplotstable,tabularx,booktabs}

\def\log{{\text{log}\;}}

\pgfplotsset{compat=1.14}
\pgfplotstableset{
  color cells/.style={
    col sep=comma,
    string type,
    postproc cell content/.code={%
            \pgfkeysalso{@cell content=\rule{0cm}{2.4ex}\cellcolor{black!##1}\pgfmathtruncatemacro\number{##1}\ifnum\number>50\color{white}\fi##1}%
            },
    columns/x/.style={
        column name={},
        postproc cell content/.code={}
    }
  }
}
\title{Dialogue Act Sequence Labeling using Hierarchical encoder with CRF}
 \author{
 	Harshit Kumar, Arvind Agarwal, Riddhiman Dasgupta, Sachindra Joshi, Arun Kumar\\
 	IBM Research, India
 }
\begin{document}
\maketitle
\begin{abstract}
Dialogue Act recognition associate dialogue acts (i.e., semantic labels) to utterances in a conversation. The problem of associating semantic labels to utterances can be treated as a sequence labeling problem. In this work, we build a hierarchical recurrent neural network using bidirectional LSTM as a base unit and the conditional random field (CRF) as the top layer to classify each utterance into its corresponding dialogue act. The hierarchical network learns representations at multiple levels, i.e., word level, utterance level, and conversation level. The conversation level representations are input to the CRF layer, which takes into account not only all previous utterances but also their dialogue acts, thus modeling the dependency among both, labels and utterances, an important consideration of natural dialogue. We validate our approach on two different benchmark data sets, Switchboard and Meeting Recorder Dialogue Act, and show performance improvement over the state-of-the-art methods by $2.2\%$ and $4.1\%$ absolute points, respectively. It is worth noting that the inter-annotator agreement on Switchboard data set is $84\%$, and our method is able to achieve the accuracy of about $79\%$ despite being trained on the noisy data. 
\end{abstract}

\section{Introduction}
\label{sec:intro}
Dialogue Acts (DA) are semantic labels attached to utterances in a conversation that serve to concisely characterize speakers' intention in producing those utterances. The identification of DAs ease the interpretation of utterances and help in understanding a conversation. One of primary applications of DAs~\cite{Higashinaka2014} is in building a natural language dialogue system, where knowing the DAs of the past utterances helps in the prediction of the DA of the current utterance, and thus, limiting the number of candidate utterances to be generated for the current turn.
For example, if the previous utterance is of type \textit{Greeting} then the next utterance is most likely going to be of the same type, i.e., \textit{Greeting}.  Table~\ref{table:daexample} shows a snippet of a conversation showing such dependency among DAs. Another application of DA identification is in building a conversation summarizer where DAs can be used to generate a summary of a conversation by collecting pair of utterances that have specific DA labels. 
\begin{table}[tbt]
  \center
  \small
  \begin{tabular}{| l | l | l | }
    \hline
     & \textbf{Utterance}& \textbf{DA} \\
    \hline
    1 & U: Hi & Greeting \\
    2 & S: Hi, How are you? & Greeting \\
    3 & U: I recently visited canary island & Statement \\
    4 & S: I am sure you had a nice time. & Statement \\
    5 & U: yes, but it is an expensive place,& Opinion \\
    6 & S: Aren\textquotesingle t all tourist places expensive? & Y/N question\\
    7 & U: yes, most are & Ack\\
    8 & U: but & Abandon \\
    9 & U: i liked the food, especially curry & Statement\\
    \hline
  \end{tabular}
  \caption{A snippet of a conversation showing few dialogues between a User (U) and System(S). }
  \label{table:daexample}
\end{table}

DA recognition is a well-understood problem, and several different approaches ranging from multi-class classification to structured prediction have been applied to it~\cite{Grau2004,Ang2005,Stolcke2006,Lendvai2007,Tavafi2013}. These approaches use handcrafted features, often designed keeping in mind the characteristics of the underlying data, and therefore do not scale well across datasets. Furthermore, in a natural conversation, there is a strong dependency among consecutive utterances, and consecutive DAs, as is evident from the previous \textit{Greeting} example, so it is important that any model should account for these dependencies. However, the standard multi-class classification such as Na\"ive Bayes does not account for any of these dependencies, and classify DAs independently, whereas structured prediction algorithms such as HMM only take into account the label dependency, not the dependencies among utterances. For the DA recognition task, one of the earlier works ~\cite{Grau2004} used Na\"ive Bayes and reported an accuracy of 66\% on the Switchboard (SwDA) corpus. The SwDA corpus has since become the standard corpus for DA recognition task because of its wide-spread use, and has been used as a benchmark data to compare different algorithms. Furthermore, structured prediction algorithms such as HMM~\cite{Stolcke2006} and SVM-HMM~\cite{Lendvai2007,Tavafi2013} though have reported an accuracy of $71\%$ and $74.32\%$, respectively, they are are still far from the human reported inter-annotator agreement of $84\%$ on SwDA corpus.

The emergence of deep learning has dramatically improved the state-of-the-art across several domains~\cite{Lecun2015}, from image classification to natural language generation. Recent studies~\cite{Blunsom2013,Lee2016,Khanpour2016,Ji2016} have used deep learning models for the DA recognition task, and have shown promising results. However, most of these models do not leverage the implicit and intrinsic dependencies among DAs. A further limitation of existing methods is that they consider a conversation as a flat structure, attempting to recognize each DA in isolation. A conversation naturally has a hierarchical structure, i.e., a conversation is made up of utterances, utterances are made up of words, and so on. In our method, we make use of this structure to build a hierarchical recurrent neural network with four layers, the first three layers representing words, utterances and conversation, and the fourth layer representing the CRF (classification) layer. Among these four layers, the first three layers capture the dependencies among utterances, whereas the fourth layer captures the dependencies among dialogue acts, hence accounting for both kind of dependencies. Our method is in contrast to the existing methods which only capture one kind of dependency either utterance dependency~\cite{Blunsom2013} or label dependency~\cite{Huang2015,Ma2016}.\\ 
The main contributions of this paper are as follows:
\begin{itemize}
\item We propose a Hierarchical Bi-LSTM-CRF (Bi-directional Long Short Term Memory with CRF) model for the DA recognition task, that can capture both kind of dependencies, i.e., among dialogue acts and among utterances.
\item We evaluate the proposed method on two benchmark datasets, SwDA and MRDA, and show performance improvement over the state-of-the-art by a significant margin. For the SwDA dataset, our method is able to achieve an accuracy of $79.2\%$ compared to the state-of-the-art accuracy of $77\%$, a step closer to the human reported inter-annotator agreement of $84\%$. On MRDA, our method achieves an accuracy of $90.9\%$ compared to the state-of-the-art accuracy of $86.8\%$.
\item We analyze the effect of incorporating linguistic features, and additional context through intra-attention~\cite{paulus2017deep} on the top of the proposed model, however, these additional variations do not result in any performance improvement. Although additional context does not boost the performance, it does help in convergence of the model at the time of training. 
\end{itemize}
\section{Related Work}
\label{sec:related}
DA recognition is a supervised classification problem that assigns DA label to each utterance in a conversation.
There exist several approaches tackling this problem in different ways, and most of them can be grouped into the following two categories: 1) those that predict the entire DA sequence for all utterances in a conversation, in other words, those that treat DA identification as a sequence labeling problem~\cite{Stolcke2006,Lendvai2007,Zimmermann2009,Lee2016}; 2) those that predict DA label for each utterance independently~\cite{Tavafi2013,Khanpour2016,Ji2016}. Until deep learning based models, the best reported accuracy on the benchmark SwDA dataset was $71\%$ by HMM~\cite{Stolcke2006}, using hand-crafted features along with contextual and lexical information, while the same for the MRDA dataset was $82\%$ by~\cite{Lendvai2007} using a naive Bayesian formulation.

Recently, researchers have started using deep learning based models for this task~\cite{Lee2016,Khanpour2016,Tavafi2013}, and have shown significant improvements over previous models. ~\cite{Lee2016} proposes a model based on CNNs and RNNs that incorporates preceding short texts as context to classify current DAs; the CNN based model performs better than the RNN based model for both SwDA and MRDA data sets. In another work, ~\cite{Blunsom2013} builds a sentence representation using a combination of Hierarchical CNN (HCNN) and RNN, followed by the classification of these sentence representation into corresponding DAs. However,~\cite{Blunsom2013} predict the dialogue act of each utterance individually, i.e., they do not take into account the label dependency. In another line of work~\cite{Ji2016}, authors propose a Latent Variable Recurrent Neural Network (LVRNN) where they tackle the problem of dialogue act classification and dialogue generation simultaneously. They use the context vector of previous utterance to predict the DA label of the next utterance which is then, along with the previous utterance vector, used to generate the next utterance. Although this model take into account the utterance dependency, it does not capture the dependencies among labels directly.

There has been some work on using conditional random fields with LSTM models~\cite{Huang2015,Ma2016} for sequence tagging tasks such as POS tagging and named entity recognition. However, they do not make use of the hierarchical structure of language, and therefore, although they take into account the label dependency, they are unable to capture the dependencies among utterances in a principled way. 


\section{Methodology}
\label{sec:method}
\begin{figure*}[!htb]
	\begin{center}
      \includegraphics[width=\textwidth]{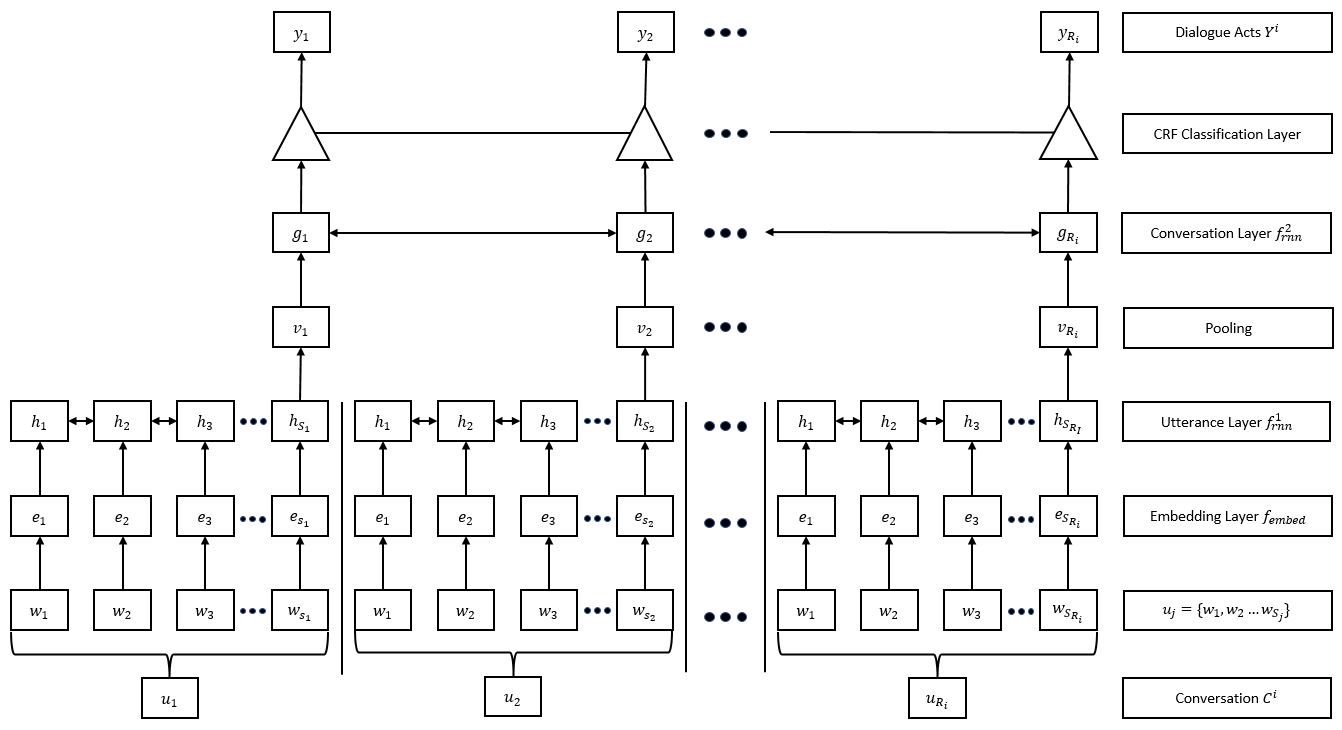}
      \caption{\small An illustration of our proposed hierarchical Bi-LSTM CRF model. The input is a conversation $C^i$ consisting of $R_i$ utterances $u_1, u_2, \dots u_{R_i}$, with each utterance $u_j$ itself being a sequence of words $w_1, w_2, \dots w_{S_j}$. As can be seen, there are four main layers, viz. embedding, utterance encoder, conversation encoder, and CRF classifier. The output is a DA prediction for each utterance in the conversation.}
    \label{fig:model}
    \end{center}
\end{figure*}
Before describing the proposed model in detail, we first set the mathematical notation for the problem of DA identification. Suppose, we have a set $\mathcal{D}$ of $N$ conversations or dialogues, i.e. $\mathcal{D} = (C^1, C^2, \dots C^N)$ with $(Y^1, Y^2, \dots Y^N)$ corresponding target DAs. Each conversation $C^i$ itself is a sequence of $R_i$ utterances $C^i = (u_1, u_2, \dots u_{R_i})$ with $Y^i = (y_1, y_2, \dots y_{R_i})$ being the corresponding target DAs. In other words, for each utterance $u_j$ in each conversation, we have an associated target label $y_j \in \mathcal{Y}$, where $\mathcal{Y}$ is the set of all possible DAs. Each utterance $u_j$ in turn is itself a sequence of $S_j$ words stringed together, i.e., $u_j = (w_1, w_2, \dots w_{S_j})$. 

The whole sequence of utterances in each conversation can be considered as a single very long chain of words, with output tags or labels only appearing sparsely, i.e., at the end of each utterance. However, such a construct suffers because of extremely long sequence lengths, which severely hampers neural network training as backpropagation through time becomes impractical due to vanishing/exploding gradients at extreme lengths. To mitigate the aforementioned problem, we take into consideration the hierarchical nature of dialogues and conversations, and opt to use a hierarchical recurrent encoder. Hierarchical recurrent encoders have been used previously by~\cite{Sordoni2015,Serban2016,Serban2017,Dehghani2017}, and have been shown to perform better compared to standard non-hierarchical models. We propose a hierarchical recurrent encoder, where the first encoder operates at the utterance level, encoding each word in each utterance, and the second encoder operates at the conversation level, encoding each utterance in the conversation, based on the representations of the previous encoder. These two encoders make sure that the output of the second encoder capture the dependencies among utterances. 

The output of the second encoder can be followed by any type of classification module which takes in the representation of each utterance, and in our formulation, we combine the hierarchical encoder with a linear chain conditional random field (CRF)~\cite{Lafferty2001} for structured prediction. DA identification can be treated as a sequence labeling problem and can be tackled naively by assigning a label to each element of the sequence independently. However, the implicit nature of dependencies among consecutive elements in a sequence means that instead of labeling each item independently, structured prediction models such as hidden Markov models, conditional random fields, etc., are naturally better choice. An illustration of the complete proposed model ---a combination of word embedding layer, a recurrent \textit{hierarchical} encoder, and a CRF based classification layer---  is shown in figure~\ref{fig:model}. The proposed model is trainable end-to-end, and constructs and captures the representation at multiple levels of granularity, e.g. word level, utterance level, and conversation level.
\subsection{Hierarchical Recurrent Encoder}
For a given conversation, each word $w_k$ of each utterance $u_j$ is processed by an embedding layer which converts one-hot vocabulary vectors to dense representations, followed by a word-level bidirectional LSTM~\cite{Hochreiter1997}, which serves as the first encoder in our hierarchical encoder. The embedding layer can be initialized using pretrained embeddings such as Word2Vec~\cite{Mikolov2013} or Glove~\cite{Pennington2014}. Since we consider bidirectional LSTMs, the representation of each word is obtained by concatenating the outputs from the forward and backward RNNs at that time-step. For an utterance $u_j$ comprised of a sequence of words $w_1, w_2, \dots w_{S_j}$, the series of operations is as follows:
\begin{align}
	\begin{split}
		e_k &= f_{embed}(w_k) \quad \forall k \in 1, 2, \dots S_j \\
    	h_k &= f^1_{rnn}(h_{k-1}, e_k) \quad \forall k \in 1, 2, \dots S_j
    \end{split}
    \label{eq:hk}
\end{align}
Here, $f_{embed}$ represents the embedding layer, whereas $f^1_{rnn}$ denotes the utterance-level encoder in our hierarchical encoder. Note that the embedding layer can ideally capture finer granularities, such as character level~\cite{Kim2016} or subword level~\cite{Sennrich2016} embeddings, which would potentially increase the depth of our hierarchical encoder. In order to keep the complexity of the model manageable, we decide to skip additional finer grained levels.

Due to the hierarchical nature of conversations, the representation of each utterance $u_j$, denoted by $v_j$ can be obtained by combining the representations of its constituent words. The combination can be done in many possible ways, e.g. average-pooling, max-pooling, etc. In the case of last pooling, we simply take the last representation of the last time-step of the word-level encoder as the representation of the entire utterance, i.e. 
\begin{equation}
	v_j =  h_{S_j}
    \label{eq:vj}
\end{equation}
This is because the final time-step contains context of all the words and time-steps preceding it, and serves as a good approximation to a representation of the entire utterance. At this stage, we have a sequence of utterance representations $v_1, v_2, \dots v_{R_i}$, corresponding to the conversation $C^i$ consisting of utterances $u_1, u_2, \dots u_{R_i}$. This sequence of utterance representation is then passed on to the conversation-level encoder which is realized by means of another bidirectional LSTM. Once again, we concatenate the vectors obtained from the forward and backward RNNs at each time-step to form the final representation of each utterance. For each utterance $u_j$, the representation $v_j$ is transformed via the utterance level encoder to obtain another representation $g_j$ as follows:
\begin{equation}
	g_j = f^2_{rnn} (g_{j-1}, v_j) \quad \forall j \in 1,2 \ldots R_i
    \label{eq:gj}
\end{equation}
Here, $f^2_{rnn}$ denotes the utterance level RNN that forms the second level in our hierarchical encoder. For a conversation $C^i$, we are left with a representation $g_j$ for each utterance $u_j$, which can be passed forward to a classification layer.
\subsection{Linear Chain CRF}
In our proposed model, the classifier of choice is a linear chain CRF, which enables us to model dependencies among labels. Note that the dependencies among utterances has already been captured by the bidirectional encoders. In sequence tagging, greedily predicting the tag at each time-step might not lead to the optimal solution, and instead, it is better to look at correlations between labels in neighborhoods in order to jointly decode the best chain of tags. CRFs are undirected graphical models that model the conditional probability of a label sequence given an observed example sequence. Now, for a given conversation $C^i$, with utterances $u_1, u_2, \dots u_{R_i}$ and corresponding associated dialogue acts $y_1, y_2, \dots y_{R_i}$, the probability of predicting the sequence of dialogue acts can be written as:
\begin{align}
	& p(y_1, y_2, \ldots y_{R_i},  u_1, u_2, \ldots u_{R_i}; \theta) = \nonumber \\ 
    & \qquad \frac
    			{\prod_{j=1}^{R_i} \psi(y_{j-1}, y_j, g_j;\theta)}
                {\sum_{\mathcal{Y}} \prod_{j=1}^{R_i} \psi(y_{j-1}, y_j, g_j;\theta)}
\end{align}
where $g_j$ is the dense representation of each utterance $u_j$ obtained from the second level encoder. Here $\theta$ is the set of parameters corresponding to the CRF layer, and $\psi()$ is the feature function, providing us with unary and pairwise potentials. The CRF layer in our proposed model is parameterized by a state transition matrix, to model the transition from a label $j-1$ to a label $j$ at any time-step. The state transition matrix is of size $K \times K$, for a tag-set of size $K$ and is position independent, i.e. it remains the same for each pair of consecutive time-steps. The transition matrix provides us with the pairwise feature function for the CRF, while the output of the hierarchical encoder, i.e. $g_j$ is considered as the unary feature function. We do not opt for higher order potentials, and restrict ourselves to only pairwise potentials, since the target sequence is a chain of tags.

To learn the CRF parameters, we use maximum likelihood training estimation. For the given training set $\mathcal{D}$, i.e. $(C^i,Y^i)$ pairs, the log likelihood can be written as:
\begin{equation}
	\mathcal{L} = \sum_{i=1}^N \log p(Y^i|C^i, \Theta)
\end{equation}
where $\Theta$ is the set of network parameters i.e. parameters of all layers, viz. word embedding layer, hierarchical recurrent encoders, and CRF classifier. At the time of testing, dynamic programming techniques~\cite{Rabiner1989} can be used to obtain the optimal sequence via the Viterbi algorithm~\cite{Viterbi1967}, i.e., 
\begin{equation}
	Y^* = \arg \max_{Y\in \mathcal{Y}} p(Y|C,\Theta)
\end{equation}

\section{Experiments}
\label{sec:exp}
In this section we describe the experimental evaluation of our approach.
\begin{table}[!htb]
  \small
  \begin{tabular}{|p{0.9cm}|p{0.3cm}|p{0.5cm}|l|l|l|}
    \hline
    Dataset & $|C|$ & $|V|$ & Training 	& Validation & Testing	\\\hline
    MRDA	& 5		& 10K	& 51(76K)	& 11(15K)	 & 11(15K) 	\\
    SwDA	& 42	& 19K	& 1003(173K)& 112(22K)	 & 19(4K) 	\\
    \hline
  \end{tabular}
  \caption{$|C|$ is the number of Dialogue Act classes, $|V|$ is the vocabulary size. Training, Validation and Testing indicate the number of conversations (number of utterances) in the respective splits.}
  \label{table:datastats}
\end{table}
\subsection{Datasets}
We evaluate the performance of our model on two benchmark datasets used in several prior studies for the DA identification task, viz.:
\begin{itemize}
  \item SwDA: Switchboard Dialogue Act Corpus~\cite{Jurafsky1997} is annotated on 1155 human to human telephonic conversations. Each utterance in a conversation is labeled with one of the 42-class compact DAMSL taxonomy \cite{Core1997}, such as STATEMENT-OPINION, STATEMENT-NON-OPINION, BACKCHANNEL, etc. 
  \item MRDA: The ICSI Meeting Recorder Dialogue Act corpus~\cite{Janin2003,Ang2005} contains 72 hours of naturally occurring multi-party meetings that were first converted into 75 word level conversations, and then hand annotated with DAs using the Meeting Recorder Dialogue Act Tagset. The original MRDA tag set had 11 general tags and 39 specific tags. The MRDA scheme provides several class-maps and corresponding scripts for grouping several related tags together into smaller number of DAs. For this work, we use the most widely used class-map that groups all tags into 5 DAs, i.e., statements (S), questions(Q), Floorgrabber (F), Backchannel (B), Disruption (D).  
\end{itemize} 
Table~\ref{table:datastats} presents different statistics for both datasets. For SwDA, train and test sets are provided but not the validation set, so we use the standard practice of taking a part of training data set as validation set~\cite{Lee2016}. Because of the noise and informal nature of utterances, we performed a series of pre-processing steps. For both datasets, exclamations and commas were stripped, and characters were converted to lower-case. The datasets are also highly imbalanced in terms of label distribution: the DA labels non-opinion (sd) and backchannel (b) in SwDA are assigned to more than $50\%$ of utterances, while more than $50\%$ of utterances in MRDA have DA label statement (s).
\begin{table}[!htb]
  \center
  \small
  \begin{tabular}{| l | l | l | }
    \hline
    \textbf{Parameter} 		& \textbf{Range}	& \textbf{Final}	\\
    \hline
    Pooling 				& Last / Mean 		& Last				\\
    Word Embedding 			& Glove / Word2Vec	& 300D Glove 		\\
    Dropout 				& $0-0.8$ 			& $0.2$ 			\\
    Bidirectional 			& True / False 		& True 				\\
    Hidden Size				& $50-300$ 			& $300$ 			\\
    Learning Rate 			& $0.5-3$ 			& $1.0$				\\
    Stacked LSTM Layers		& $1-4$ 			& $1$ 				\\
    \hline
  \end{tabular}
  \caption{Hyperparameter tuning -- the $2^{\text{nd}}$ column lists the various values tried, while the $3^{\text{rd}}$ column lists the final value chosen for the corresponding hyperparameter.}
  \label{table:parameter}
\end{table}
\subsection{Hyperparameter Tuning}
Conversations with the same number of utterances were grouped together into mini-batches, and each utterance in a mini-batch was padded to the maximum length for that batch. The maximum batch-size allowed was $64$. We used $L2$ regularization of $1e-4$ in the form of weight decay and the Adadelta optimizer. All other hyper-parameters were selected by tuning one hyper-parameter at a time while keeping the others fixed. The hyper-parameters were tuned using the SwDA validation set. The final set of hyper-parameters were then used to train two different models, one each on SwDA and MRDA training datasets. Table~\ref{table:parameter} lists the range of values for each parameter that we experimented with, and the final value that was selected. The word vectors were initialized with the 300-dimensional Glove embeddings~\cite{Pennington2014}, and were also updated during training. Dropout was applied to the embeddings obtained from the output of each encoder. The learning rate was initialized to $1.0$ and reduced by a factor of $0.5$ every $5$ epochs. Early stopping is also used on the validation set with a patience of $5$ epochs. Increasing the number of stacked LSTM layers reduced the accuracy of the model, so we settled with only one layer.
\subsection{Results and Discussion}
The results reported in this section are based on the hyper-parameters values tuned in the previous section. The Hierarchical Bi-LSTM-CRF model is compared against seven different baseline models. 
\begin{itemize}
  \item \textbf{DRLM-Conditional}~\cite{Ji2016} - a latent variable recurrent neural network architecture for joint modeling of utterance and DA label. 
  \item \textbf{LSTM-Softmax}~\cite{Khanpour2016} - Bidirectional LSTMs on word embeddings followed by a softmax classifier. 
  \item \textbf{RCNN}\cite{Blunsom2013} - Hierarchical CNN on word embeddings to model utterances followed by a RNN to capture context, with a softmax classifier.
  \item \textbf{CNN}\cite{Lee2016} - An utterance level CNN followed by a conversation CNN, with softmax classifiers. The utterance and conversation layers only consider the current utterance and at most $2$ preceding ones.
  \item \textbf{CRF} - Simple baseline with pre-trained word embeddings followed by a CRF classifier.
  \item \textbf{LR} - Simple baseline with pre-trained word embeddings followed by a logistic regression classifier.
\end{itemize}
\begin{table}[!htb]
  \center
  \small
  \begin{tabular}{| l | l |}
    \hline
    \textbf{Model} 						& \textbf{Acc(\%)} 	\\
    \hline
    Hierarchical Bi-LSTM-CRF 			& \textbf{79.2} 	\\
    DRLM-Conditional(Ji et al. 2016) 	& 77.0 				\\
    LSTM-Softmax(Khanpour et al. 2016)	& 75.8\footnotemark	\\
    RCNN\cite{Blunsom2013} 				& 73.9 				\\
    CNN\cite{Lee2016} 					& 73.1 				\\
    CRF 								& 72.2				\\
    LR 									& 71.4				\\
    HMM\cite{Stolcke2006} 				& 71.0 				\\
    \hline
  \end{tabular}
  \caption{Comparing accuracy of our method (Hierarchical Bi-LSTM-CRF) with other methods in the literature on SwDA dataset. }
  \label{table:results}
\end{table}
\footnotetext{The paper claimed accuracy of 80.1. Personal correspondence with the authors revealed that a non-standard test set was used by accident.}
Table~\ref{table:results} compares the results obtained using our model with the other previous models. The results show that our Hierarchical Bi-LSTM-CRF model outperforms the state-of-the-art. Our model improved the DA labeling accuracy over DRLM-Conditional model by $2.2\%$ absolute points. In order to further analyze the results, we looked into the confusion matrix to know which labels are incorrectly/correctly assigned to utterances. Table~\ref{table:swdaconfusion} shows the confusion matrix of our proposed model for the SwDA dataset. Among them the most confused pairs are (sd,sv) and (aa,b) which represent (statement-non-opinion, statement-opinion) and (agree-accept, acknowledge) respectively. The total number of utterances with DA 'sd', 'sv', 'aa', and 'b' are $1317$, $717$, $208$, and $762$, respectively. $103$ utterances (7.8\%) with true label \textit{non-opinion} were predicted incorrectly as \textit{opinion}, whereas, $1155$ utterances (87.7\%) with true label \textit{non-opinion} were predicted correctly. Similarly, $200$ utterances (27.9\%) with true label \textit{opinion} were predicted incorrectly as \textit{non-opinion} whereas $473$ utterances (66\%) with true label \textit{opinion} were predicted correctly. On further analysis of the cause of this confusion between these two class pairs, we identified that there are utterances which were classified correctly by the model, however, they were marked incorrectly classified because of bias in the ground truth. For some of the utterances, classes were not distinguishable even by humans because of the subjectivity.
\begin{table}[h]
	\setlength{\tabcolsep}{1pt}
	\centering
	\small
	\pgfplotstableset{
    	color cells/.style={
        col sep=comma,
        string type,
        postproc cell content/.code={%
        	\pgfkeysalso{@cell content=\rule{0cm}		{2.4ex}\cellcolor{black!##1}\pgfmathtruncatemacro\number{##1}\ifnum\number>50\color{white}\fi##1}%
                },
	        columns/a/.style={
    	        column name={},
        	    postproc cell content/.code={},
        	},
        	columns/qyd/.style={
            	column name={qy\^d},
            	postproc cell content/.code={}
        	},
         	columns/p/.style={
            	column name={\%},
            	postproc cell content/.code={}
        	}
    	}
  	}
    \pgfplotstabletypeset[color cells]{
    a,fo,qw,qyd,qy,sd,ad,h,aa,b,sv
    (16) fo, 62.5,6.3,0.0,0.0,6.3,0.0,0.0,0.0,0.0,0.0
    (55) qw, 0.0,78.2,0.0,0.0,3.6,1.8,0.0,0.0,0.0,0.0
    (36) qy\^d, 0.0,0.0,19.4,27.8,30.6,0.0,0.0,0.0,0.0,16.7
    (84) qy, 0.0,1.2,1.2,79.8,2.4,0.0,0.0,0.0,0.0,3.6
    (1317) sd, 0.0,0.0,0.6,0.0,87.7,0.2,0.1,0.5,0.2,7.8
    (27) ad, 0.0,0.0,0.0,0.0,33.3,29.6,0.0,0.0,0.0,33.3
    (23) h, 0.0,0.0,0.0,0.0,17.4,0.0,60.9,0.0,0.0,8.7
    (208) aa, 0.0,0.0,0.0,0.5,2.9,0.0,0.0,75.5,16.4,1.9
    (762) b, 0.0,0.0,0.0,0.0,0.0,0.0,0.0,5.9,88.6,0.0
    (717) sv, 0.0,0.0,0.3,0.1,27.9,0.3,0.1,1.0,0.0,66.0
	}
	\caption{Confusion matrix of Hierarchical Bi-LSTM-CRF model for the SwDA dataset (10 DA class labels), where the row denotes the true label and the column denotes the predicted label. The numbers in the bracket besides the DA label in the first cell of each row is the count of the number of utterances of that DA label.}
  	\label{table:swdaconfusion}
\end{table}
\begin{table*}[bp]
  \centering
  \small
  \begin{tabular}{|l|l|l|l|}
    \hline
    Utt no & Utterance                               & True DA Label 		& Predicted DA Label	\\\hline
    1692   & This is quite a long distance.          & non-opinion (sd)		& opinion (sv) 			\\
    1720   & This is a little bigger than a tea cup. & non-opinion (sd)		& opinion (sv)			\\
    1789   & we're supposed to appreciate them.      & non-opinion (sd)		& opinion (sv)			\\\hline
    77     & they could do something about that      & opinion (sv)			& non-opinion (sd)		\\
    739    & i need to start jog something again.    & opinion (sv)			& non-opinion (sd)		\\
    112    & i thought it was up there.              & opinion (sv)			& non-opinion (sd)		\\\hline
    1121   & Yeah.                                   & agree/accept (aa)	& backchannel (b)		\\
    1334   & Yeah.                                   & agree/accept (aa)	& backchannel (b)		\\
    1337   & Sure                                    & agree/accept (aa)	& backchannel (b)		\\\hline
    1362   & Yeah                                    & backchannel (b)		& agree/accept (aa)		\\
    1371   & Yeah                                    & backchannel (b)		& agree/accept (aa)		\\
    1372   & \# Oh Yeah. \#                          & backchannel (b)		& agree/accept (aa)		\\\hline
  \end{tabular}
  \caption{Example of utterances of confused pairs (non-opinion, opinion) and (agree/accept, backchannel)}
  \label{table:confusedpairs}
\end{table*}

We show examples of some of these cases in Table~\ref{table:confusedpairs}. For instance, the utterance no. 1692 seems to be an opinion ('sv') and is also predicted as 'sv', but its true label is non-opinion ('sd'). Similarly, utterance no. 1334 underlying text is 'Yeah', its true label is agree/accept ('aa'). Also, utterance no. 1362 and 1371 underlying text is 'Yeah', this time its true label is backchannel('b'). This means two utterances with the same underlying text have two different DA associations. 
We accepted it as the characteristics of the SwDA dataset, this thought is echoed by the authors who created the dataset that the inter-labeler agreement is $84.0\%$. 
\begin{table}[!htb]
  \center
  \small
  \begin{tabular}{| l | l |}
    \hline
    \textbf{Model} & \textbf{Acc(\%)} \\
    \hline
    Hierarchical Bi-LSTM-CRF & \textbf{90.9} \\
    LSTM-Softmax(Khanpour et al. 2016) & 86.8\\
    CNN\cite{Lee2016} & 84.6 \\
    Naiive Bayes\cite{Lendvai2007} & 82.0 \\
    \hline
  \end{tabular}
  \caption{Comparing Accuracy of our method (Bi-LSTM-CRF) with other methods in the literature on the MRDA dataset.}
  \label{table:results1}
\end{table}

The results on the MRDA dataset are shown in Table~\ref{table:results1}. From this table, it is clear that our method outperforms the state-of-the-art by a significant margin i.e. by 4.1\%. Table~\ref{table:mrdaconfusion} shows the confusion matrix for the MRDA dataset. Except for the class label 'B', all other DA class labels are predicted accurately. Approximately 21\% of DA class label 'B' are incorrectly predicted as 'S'. One of the reasons for this behavior is that the MRDA dataset is highly imbalanced, with more than 50\% of the utterances labeled as class 'S'.
\begin{table}[!htb]
  \centering
  \small
  \pgfplotstabletypeset[color cells]{
       x,F,D,S,B,Q
  (1314) F,80.06,4.95,6.7,8.3,0.00
  (2244) D,4.86,90.06,2.45,2.41,0.22
  (8564) S,0.39,0.02,94.69,4.85,0.06
  (1961) B,1.12,0.15,20.96,77.77,0.00
  (1112) Q,0.00,0.00,0.54,0.00,99.46
   }
  \caption{Confusion matrix of Bi-LSTM-CRF for the MRDA dataset, where the row denotes the true DA label and the column denotes the predicted DA label. The numbers in the bracket besides the DA label in the first cell of each row is the count of the number of utterances of that DA label.}
  \label{table:mrdaconfusion}
\end{table}
\subsection{Effect of Hierarchy and Label Dependency}
In this section, we discuss the influence of adding hierarchical layers (utterance layer, conversation layer) and classification layer on accuracy. In particular, we perform ablation studies by evaluating the model layer by layer to understand if the addition of new layers provides any improvement in performance.

The first model, WE, is a plain two layer network with a word embedding layer followed by the classification layer, i.e., the pre-trained Glove word embeddings are fed as input to the classification layer. No form of dependency, among utterances, across utterances, across DA labels, are captured here. The second model, WE+UL, is a three layer network that takes word embeddings as input. The output of WE layer is input to the utterance layer to learn utterance vectors. Each utterance vector is a compositional representation of all words in that utterance. Utterance vector is fed as input directly to the classification layer to predict the label. Dependencies across utterances are not captured here. The third model, WE+UL+CL, is a four layer network similar to the proposed hierarchical Bi-LSTM-CRF model, except that the final layer can be either logistic regression (LR) or a CRF based classifier. 
\begin{table}[!htb]
  \centering
  \small
  \begin{tabular}{|l|l|l|}
    \hline
    \textbf{Model}	& \textbf{Accuracy} & \textbf{Accuracy} \\
                    & \textbf{with LR} 	& \textbf{with CRF}	\\\hline
    WE 				& 71.4				& 72.2 				\\
    WE+UL 			& 72.2 				& 72.7				\\
    WE+UL+CL 		& 74.1 				& 79.2				\\
    \hline
  \end{tabular}
  \caption{WE is Word Embedding layer, UL is Utterance Layer, CL is Conversation Layer, LR is Logistic regression and CRF is Conditional Random Field.}
  \label{table:journey}
\end{table}

Table~\ref{table:journey} shows the results of various networks with both LR and CRF layer. From the table, we observe that the models WE, WE+UL, and WE+UL+CL with LR layer at the top produce an accuracy of $71.4\%$, $72.2\%$, and $74.1\%$, respectively. In the final layer, if LR is replaced with CRF then the accuracy of WE, WE+UL, and WE+UL+CL (Hierarchical Bi-LSTM-CRF) is $72.2\%$, $72.7\%$, and $79.2\%$, respectively. From these results it is clear that adding additional layers, viz. utterance layer and conversation layer, improve the results by a few notches. Also, replacing LR with CRF further improves the results. Note that the accuracy of WE+UL with LR and WE with CRF is same. We understand that the output of utterance layer at each time step is a vector representing the context of the utterance till that word. The word vector at the last time step is the final representation of the utterance. This means, adding an utterance layer generates a compositional vector of all words in an utterance, and thus serves as a good representation of all words in the utterance. Adding the utterance layer and replacing the LR with CRF in the existing model produces more or less the same result. Addition of conversation layer results in major improvement in the accuracy, approximately $2\%$ absolute points with LR in the final layer, and $6\%$ absolute points with CRF . This is because the output of conversation layer for an utterance is a representational vector capturing the context of itself and utterances preceding it. 
\subsection{Effect of Linguistic Features and Context}
For Dialogue Act identification, linguistic features-\cite{Tavafi2013} and context information~\cite{ribeiro2015influence} have shown to improve the performance of the underlying model. In our model, we add linguistic features, in particular the part-of-speech tags (POS) associated with words in an utterance. More specifically, we add a POS tag layer with POS tag embeddings followed by an encoder, working in parallel to the utterance encoder, to learn a representation for each POS tag sequence associated with each utterance, and concatenate it with the utterance vector at the conversation layer, right before they are fed to the CRF layer. The results show that the addition of POS reduces the accuracy by approximately $1\%$.
\begin{table}[h]
  \centering
  \small
  \begin{tabular}{@{}|c|l|l|@{}}
    \hline
    \multicolumn{2}{|c|}{\textbf{Extension}} 	& \textbf{Accuracy(\%)} \\\hline
    \multicolumn{2}{|l|}{POS} 								& 77.9 		\\ \hline
    \multirow{3}{*}{Context} 					& length 10 & 77.4 		\\
     											& length 5 	& 78.3 		\\
     											& length 3 	& 78.1 		\\ \hline
  \end{tabular}
  \caption{Accuracy obtained using two extensions to the Hierarchical Bi-LSTM-CRF model.}
  \label{table:extension}
\end{table}

In another extension, we explore capturing context of an utterance through intra-attention~\cite{paulus2017deep}, and concatenating it to the utterance vector to produce a new utterance vector. Recent research~\cite{Cho2014} has shown that LSTM performance deteriorates as the length of input sentence increases since they are not able to capture long context. Therefore, capturing context explicitly through attention~\cite{Bahdanau2015} is an alternate way to model long-term dependencies. In our model, after obtaining utterance vectors from the conversation layer, a normalized attention weight vector is computed for each utterance vector, by computing its similarity from previous utterance vectors. These attention weights are then used to compute the context vector by taking a weighted sum of the previous $K$ utterance vectors. The new context vector is concatenated to the utterance vector produced by the conversation layer to obtain new utterance vector, which is input to the classification layer. We experimented with this attention by varying the length of the context (number of previous utterances) i.e. $K \in (10, 5, 3)$.  In a conversation, an utterance at time step $t$ is mostly dependent upon the previous two or three utterances. Modeling too long dependencies therefore reduces the performance, as is shown in Table~\ref{table:extension}. 

Overall, adding additional context or POS representations to the Hierarchical Bi-LSTM-CRF model does not improve the performance, which means, these new additions are not contributing any new information to the existing model. The original hierarchical encoder has all the required information it needs to model the utterance representation and the dependencies among them. Although additional context does not help in performance, it helps quite a bit in convergence. We observed that training the model with additional context results in much faster convergence compared to training without context. For the SwDA dataset, the accuracy with additional context and without it after the first epoch was $68.8\%$ and $65.1\%$, respectively. Similarly, for the MRDA dataset, the accuracy after first epoch while training the model with additional context was $88\%$, whereas without it was $87\%$.

\section{CONCLUSION}
\label{sec:conclusion}
In this paper, we used a Hierarchical Bi-LSTM-CRF model for labeling sequence of utterances in a conversation with Dialogue Acts. The proposed model captures long term dependencies between words in an utterance and across utterances, thus generating vector representations for each utterance in a conversation. The sequence of vectors corresponding to utterances in a conversation are sent to a CRF based classifier to model the dependencies between the Dialog Act labels and the utterance representations. We demonstrated the efficacy of our model on two popular datasets, SwDA and MRDA. Experimental results highlight that our proposed model outperforms the state-of-the-art for both data sets.
\fontsize{9.5pt}{10.5pt} \selectfont
\bibliographystyle{aaai}
\bibliography{strings}
\end{document}